\documentclass[10pt,twocolumn,letterpaper]{article}

\usepackage{cvpr} 

\definecolor{cvprblue}{rgb}{0.21,0.49,0.74}
\usepackage[pagebackref,breaklinks,colorlinks,allcolors=cvprblue]{hyperref}
\usepackage{graphics,graphicx,float,subcaption,booktabs,xcolor,multirow,array,color,ifthen,tabu,colortbl,dblfloatfix,url,relsize,algorithm,amssymb,xspace,nicefrac,microtype,amsmath,amsfonts,bm,wrapfig,makecell,enumitem}
\usepackage[font=small]{caption}
\usepackage[capitalize]{cleveref}
\usepackage[english]{babel}
\usepackage[accsupp]{axessibility}

\newcommand{\method}{CoT-VLA}

\usepackage{algorithm}
\usepackage{algpseudocode}

\def\paperID{10930} 
\def\confName{CVPR}
\def\confYear{2025}

\title{\method: Visual Chain-of-Thought Reasoning for\\Vision-Language-Action Models}

\author{Qingqing Zhao\textsuperscript{1,2,\thanks{Work done during an internship at NVIDIA.}}\;\;\; Yao Lu\textsuperscript{1}\;\;\; Moo Jin Kim\textsuperscript{2}\;\;\; Zipeng Fu\textsuperscript{2}\;\;\; \\ Zhuoyang Zhang\textsuperscript{3}\;\;\; Yecheng Wu\textsuperscript{1,3}\;\;\; Zhaoshuo Li\textsuperscript{1} \;\;\; Qianli Ma\textsuperscript{1} \;\;\; Song Han\textsuperscript{1,3} \;\;\; Chelsea Finn\textsuperscript{2} \;\;\; \\ Ankur Handa\textsuperscript{1} \;\;\; Ming-Yu Liu \;\;\; Donglai Xiang\textsuperscript{1\thanks{Equal Advising}} \;\;\; Gordon Wetzstein\textsuperscript{2$^\dagger$}   \;\;\;  Tsung-Yi Lin\textsuperscript{1$^\dagger$} \vspace{2mm}\\
\textsuperscript{1}NVIDIA\;\;\;
\textsuperscript{2}Stanford University \;\;\;  \textsuperscript{3}MIT\\
}

\begin{document}
\maketitle

\begin{abstract}
Vision-language-action models (VLAs) have shown potential in leveraging pretrained vision-language models and diverse robot demonstrations for learning generalizable sensorimotor control. While this paradigm effectively utilizes large-scale data from both robotic and non-robotic sources, current VLAs primarily focus on direct input--output mappings, lacking the intermediate reasoning steps crucial for complex manipulation tasks. As a result, existing VLAs lack temporal planning or reasoning capabilities. In this paper, we introduce a method that incorporates explicit visual chain-of-thought (CoT) reasoning into vision-language-action models (VLAs) by predicting future image frames autoregressively as visual goals before generating a short action sequence to achieve these goals. We introduce \method{}, a state-of-the-art 7B VLA that can understand and generate visual and action tokens. Our experimental results demonstrate that \method{} achieves strong performance, outperforming the state-of-the-art VLA model by 17\% in real-world manipulation tasks and 6\% in simulation benchmarks. 
Videos are available at: \url{https://cot-vla.github.io/}.
\end{abstract}

\begin{figure}[t]
    \includegraphics[width=\linewidth]{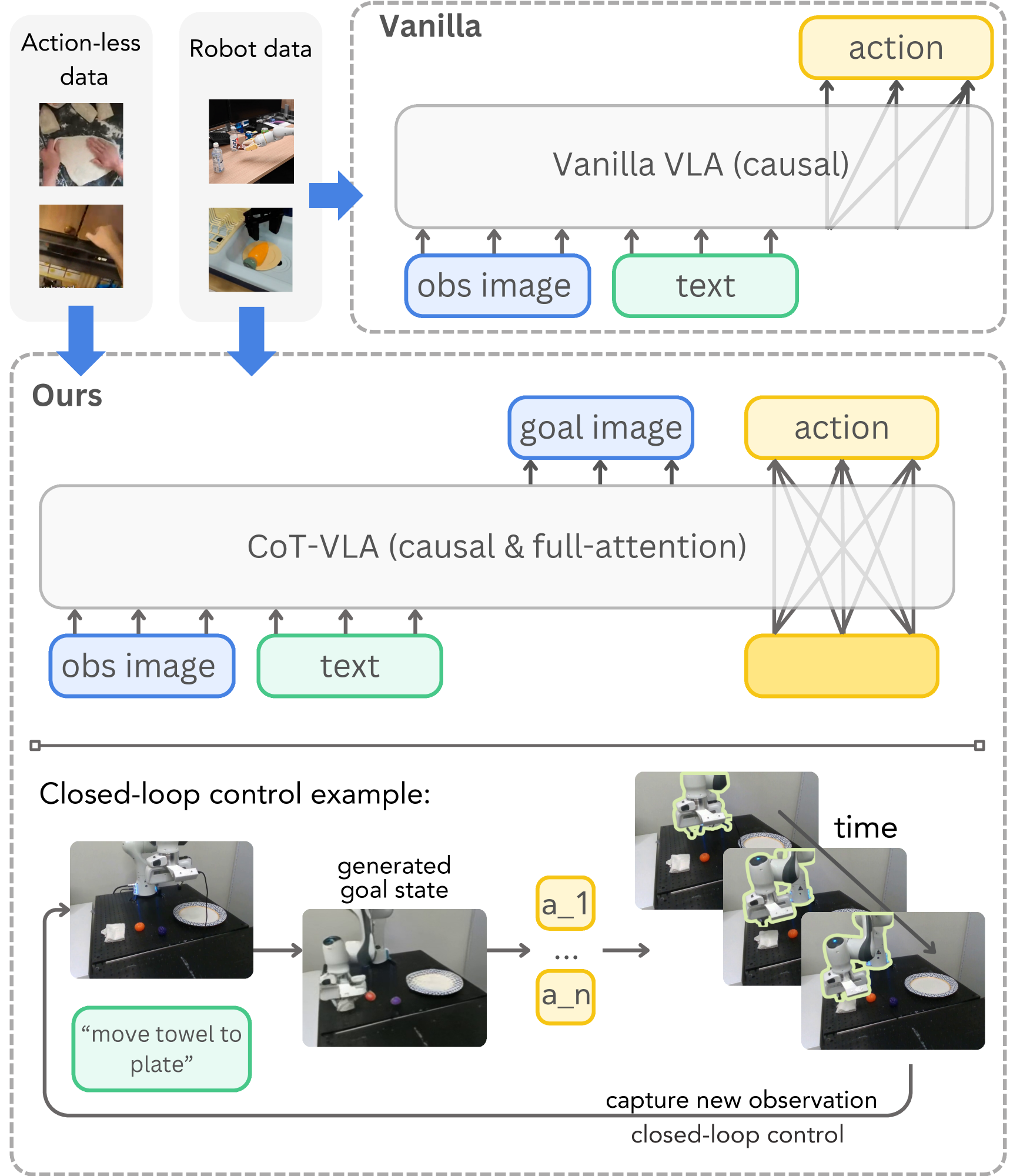}
    \caption{\textbf{Comparison between vanilla VLA and \method{} frameworks}. Prior VLA models (top) directly predict robot actions from task inputs without explicit reasoning steps and only use action-annotated robot demonstration data for training. Unlike vanilla VLAs, \method{} (bottom) can also leverage action-less datasets like EPIC-KITCHEN-100~\cite{kapidis2019egocentric} to enhance subgoal image generation ability, unlocking the potential of using abundant unlabeled video data to improve VLA's visual reasoning capability. \method{} first generates a subgoal image as an intermediate reasoning step, and then generate a short action sequence to achieve the subgoal. We outline the robot arm for better visualization.}
    \label{fig:teaser}
    \vspace{-10pt}
\end{figure}

\section{Introduction}
Recent advances in robot learning have demonstrated impressive progress in training policies that can act across diverse tasks and environments~\cite{kim2024openvla,brohan2022rt,bharadhwaj2024gen2act,du2023video,mu2024embodiedgpt,michal2024robotic,zhang2024learning,yang2024pushing,wen2023any,o2023open,team2024octo,wu2023unleashing,driess2023palm,3dvla,shridhar2022cliport,lin2024data,cheang2024gr,fu2024mobile}. One promising direction is vision-language-action (VLA) models, which leverage the rich understanding capabilities of pretrained vision-language models (VLMs) to map natural language instructions and visual observations to robot actions~\cite{driess2023palm,kim2024openvla,o2023open}. By training VLMs on robot demonstrations, VLAs inherit their ability to understand diverse scenes, objects, and natural language instructions, leading to better generalization capabilities when fine-tuned for downstream testing scenarios. While these approaches have shown impressive results, they typically map directly from observations to actions without explicit intermediate reasoning steps that could improve interpretability and, potentially, performance.

In the language domain, chain-of-thought (CoT) prompting has emerged as a powerful technique for improving the reasoning capabilities of large language models (LLMs) by encouraging step-by-step thinking~\cite{wei2022chain,zelikman2022star}. Applying these concepts to robotics presents exciting opportunities for grounding reasoning in text, visual observations, and physical actions. Recent works have made progress in this direction by incorporating intermediate reasoning steps like language descriptions, keypoints, or bounding boxes~\cite{du2024learning,michal2024robotic,mu2024embodiedgpt,wen2023any}. These intermediate representations capture abstracted states of scenes, objects, and tasks and often require additional pre-processing pipelines. In our work, we explore subgoal images as an intermediate reasoning step before action generation. These images capture the state of the model's reasoning process and are naturally available within robot demonstration datasets.
While prior work has explored subgoal generation and goal-conditioned imitation learning~\cite{susie,nair2018visual,ding2019goal,shridhar2024generative}, to the best of our knowledge, our approach is the first to integrate these concepts with VLAs as intermediate chain-of-thought reasoning steps.

We propose visual chain-of-thought reasoning for VLAs, a new method that uses subgoal image generation as a form of chain-of-thought reasoning for robotic tasks. Rather than directly predicting actions, our method first generates a subgoal image that represents the robot's planned state in pixel space, and then conditions its action on both the current observation and the generated subgoal image. This approach allows the model to ``think visually" about how to accomplish a task before acting. By using the subgoal image as intermediate reasoning step, we leverage information that already exists in robot manipulation data with minimal pre-processing required. Furthermore, since subgoal image generation does not require action annotations, this unlocks the potential of using abundant video data for improved visual reasoning and understanding.

We build our \method{} system that leverages visual chain-of-thought reasoning upon recent advances in unified multimodal foundation models that can understand and generate text and images~\cite{wu2024vila,wang2024emu3,lu2024unified,team2024chameleon,xie2024show}. We train our base model~\cite{wu2024vila} on both the Open X-Embodiment dataset~\cite{o2023open} and action-less video datasets~\cite{goyal2017something, kapidis2019egocentric}, and then fine-tune the model on task demonstrations collected on downstream robot setups used for deployment and evaluation. We design a hybrid attention mechanism for \method{}: we use causal attention with next-token prediction for text and image generation, and leverage full attention to predict all action dimensions at once. Additionally, inspired by recent advances in robot learning~\cite{zhao23act,fu2024humanplus, chi2023diffusion}, we predict sequences of actions (action chunking) rather than a single action at each timestep. We demonstrate that both action chunking and the hybrid attention mechanism improve the model's performance.

Through extensive experiments in both simulation benchmarks~\cite{liu2023libero} and real-world experiments\cite{walke2023bridgedata,o2023open}, we demonstrate that our visual chain-of-thought reasoning helps improve policy performance compared to prior VLA approaches. Our key contributions include:
\begin{itemize}
    \item We introduce a method of visual chain-of-thought reasoning through subgoal image generation as an intermediate reasoning step for robotic control.
    \item We introduce a system \method{} that incorporates visual chain-of-thought reasoning, and a hybrid attention mechanism that combines causal attention for pixel and text generation and full attention for action prediction.
    \item We conduct comprehensive evaluations in both simulation and the real world, demonstrating that visual chain-of-thought reasoning improves VLA performance, and our system achieves state-of-the-art performance across multiple robot platforms and tasks.
\end{itemize}

\section{Related Work}
\paragraph{Chain-of-Thought (CoT) Reasoning} 
CoT reasoning has gained prominence in natural language processing, particularly for enabling models to perform complex, multi-step reasoning tasks by breaking down problem-solving into sequential, explainable steps. Early work on CoT reasoning~\cite{wei2022chain} has demonstrated the effectiveness of prompting large language models to generate intermediate reasoning steps before arriving at a final answer. 
Extending this paradigm to the visual domain, researchers have explored multimodal chain-of-thought methods, where visual information is processed iteratively in a stepwise fashion to reason about future outcomes or states, including generating bounding boxes~\cite{shao2024visual}, intermediate image infillments using Stable Diffusion~\cite{rose2023visual} or standard Python packages~\cite{hu2024visual}, or generating CLIP embeddings~\cite{harvey2023visual}.
Recently, CoT reasoning has been explored in embodied applications. It can generate textual plans for multi-stage execution~\cite{mu2024embodiedgpt,michal2024robotic}, point trajectories~\cite{wen2023any}, label bounding boxes of objects and gripper positions as additional observations~\cite{michal2024robotic}, generate future image trajectories for open-loop following~\cite{ni2024generate,liang2024dreamitate}, and generate fine-grained reward guidance for reinforcement learning~\cite{zhang2024learning}.
In this work, we introduce Visual-CoT reasoning for robotic manipulation, where predicted subgoal images serve as intermediate reasoning steps for closed-loop action generation. This approach leverages demonstration videos as natural intermediate reasoning states without requiring additional annotations.

\begin{figure*}[t!]
    \centering
    \includegraphics[width=\linewidth]{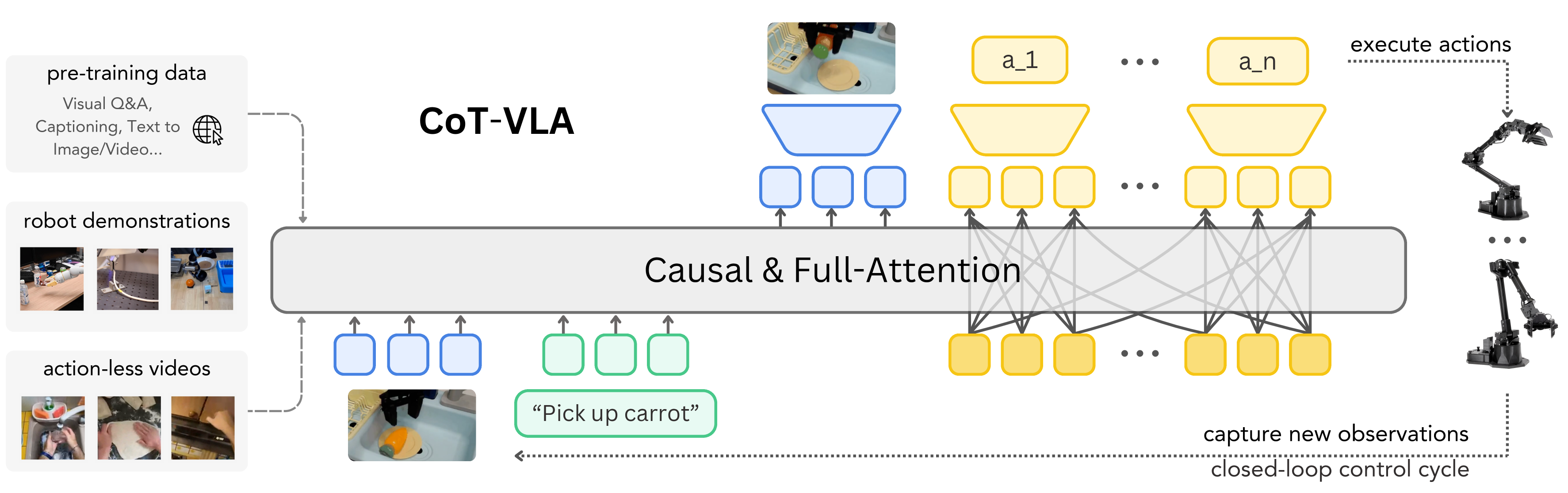}
    \caption{\textbf{Overview of \method{} framework}. We build our model on VILA-U~\cite{wu2024vila}, a generative multimodal model pretrained on interleaved text--image data. The base model then trains on robot demonstrations~\cite{o2023open} and action-less videos~\cite{goyal2017something, kapidis2019egocentric}. During deployment, given a visual observation and a text instruction, the model performs visual chain-of-thought reasoning by generating a subgoal image (upper blue) with causal attention. It then generates a short action sequence with full attention ($\mathbf{a}_1 \cdots \mathbf{a}_n$) for robot execution. The system operates in a closed-loop control manner by capturing new observations after executing predicted action sequences.}
    \label{fig:method}
\end{figure*}

\paragraph{Vision-Language-Action Models}
Large pretrained vision-language models (VLMs)~\cite{karamcheti2024prismatic,liu2024visual,chen2023pali} have emerged as a powerful tool for robot learning, and recent works have explored various approaches to integrate them into robot systems. Several works utilize VLMs as intermediate components for perception and control, leveraging their strong semantic understanding and reasoning capabilities to decompose complex tasks~\cite{huang2024rekep,li2024driving,singh2023progprompt,ha2023scaling}, detect objects~\cite{gadre2023cows,huang2023visual}, or generate dense rewards~\cite{ma2023eureka,du2023reward,yu2023language}
or goals~\cite{susie,ni2024generate,ding2019goal,nair2018visual,zhou2024minedreamer,yang2023learning,du2024learning,shridhar2024generative}. Some approaches incorporate VLMs~\cite{radford2021learning,caron2021emerging,wen2024tinyvla} into end-to-end trainable policies by using them as pretrained backbone for better visuo-language representation~\cite{kim2024openvla,o2023open,team2024octo,driess2023palm,majumdar2023we}. Most relevant to our work are recent approaches that fine-tune pretrained VLMs on robot demonstration data for direct action prediction~\cite{o2023open,kim2024openvla,driess2023palm}. These VLAs demonstrate improved generalization to novel objects, environments, and natural language instructions through pretraining on internet-scale vision-language datasets, providing a promising direction for transferring visual and language knowledge to robotic control tasks. However, most existing VLAs do not leverage the step-by-step reasoning capabilities demonstrated in large language models, which have been shown to significantly improve performance across various tasks~\cite{wei2022chain}. In the past, researchers have used chain-of-thought reasoning on language instructions or intermediate keypoints/bounding boxes for robotics~\cite{du2023reward,Chen_2024_CVPR,mu2024embodiedgpt,wen2023any}. We introduce visual chain-of-thought reasoning to the VLA frameworks, using subgoal images as intermediate reasoning steps before action generation.

\section{\method{}}
In this section, we present our visual chain-of-thought reasoning framework for VLAs. We begin with the formulation of our method (\ref{sec:cot}), followed by a detailed description of the system architecture (\ref{sec:vilau}). We then explain our training procedures (\ref{sec:trianing}) and outline the deployment strategy for downstream tasks (\ref{sec:deployment}).

\subsection{Visual Chain-of-Thought Reasoning}\label{sec:cot}
We consider two types of training data for VLA pretraining: robot demonstrations dataset $D_r$ and action-less videos dataset $D_v$. Robot demonstrations are represented as $D_r = \{(l, \mathbf{a}_{1...T}, \mathbf{s}_{1...T})\}$, where $l$ denotes the natural language instruction, $\mathbf{a}_{1...T} = \{\mathbf{a}_1,...,\mathbf{a}_T\}$ denotes the sequence of robot actions, and $\mathbf{s}_{1...T} = \{\mathbf{s}_1,...,\mathbf{s}_T\}$ denotes the visual observations as a sequence of images. Action-less videos $D_v = \{(l, \mathbf{s}_{1...T})\}$ consist of language descriptions and images without action annotations.

\paragraph{VLA:} Vanilla VLA approaches fine-tune a pretrained VLM, $P_\theta$, on $D_r$, learning to predict actions $\mathbf{\hat{a}}_{t+1}$ directly from the current observation $\mathbf{s}_t$ and language instruction $l$ (Figure~\ref{fig:teaser}, top):
\begin{align}
    \mathbf{\hat{a}}_{t} \sim P_\theta(\mathbf{a}_{t}|\mathbf{s}_t,l)
\end{align}
\paragraph{\method:} Our key insight is to incorporate explicit visual reasoning before action generation. As illustrated in Figure~\ref{fig:method}, our approach operates in two sequential phases:
\begin{align}
    \mathbf{\hat{s}}_{t+n} \sim P_\theta(\mathbf{s}_{t+n}|\mathbf{s}_t,l) \label{eq:cot} \\
    \{\mathbf{\hat{a}}_t, ..., \mathbf{\hat{a}}_{t+m}\} \sim P_\theta(\{\mathbf{a}_t, ..., \mathbf{a}_{t+m}|\mathbf{s}_t,l,&\mathbf{{s}}_{t+n}) \label{eq:cotvla}
\end{align}
where we first predict a subgoal image $\mathbf{\hat{s}}_{t+n}$, $n$ frames ahead, as an intermediate visual reasoning step (Equation~\ref{eq:cot}). Then we generates a sequence of $m$ actions to achieve this subgoal state (Equation~\ref{eq:cotvla}). This enables the model to ``think visually" first by explicitly reasoning about desired future states before predicting the actions. The visual reasoning step, Equation \eqref{eq:cot}, is trained on both robot demonstrations $D_r$ and action-less videos $D_v$, and the action generation step, Equation \eqref{eq:cotvla}, is trained on robot demonstrations $D_r$ only.

\begin{figure}[h]
    \centering
    \includegraphics[width=0.75\linewidth]{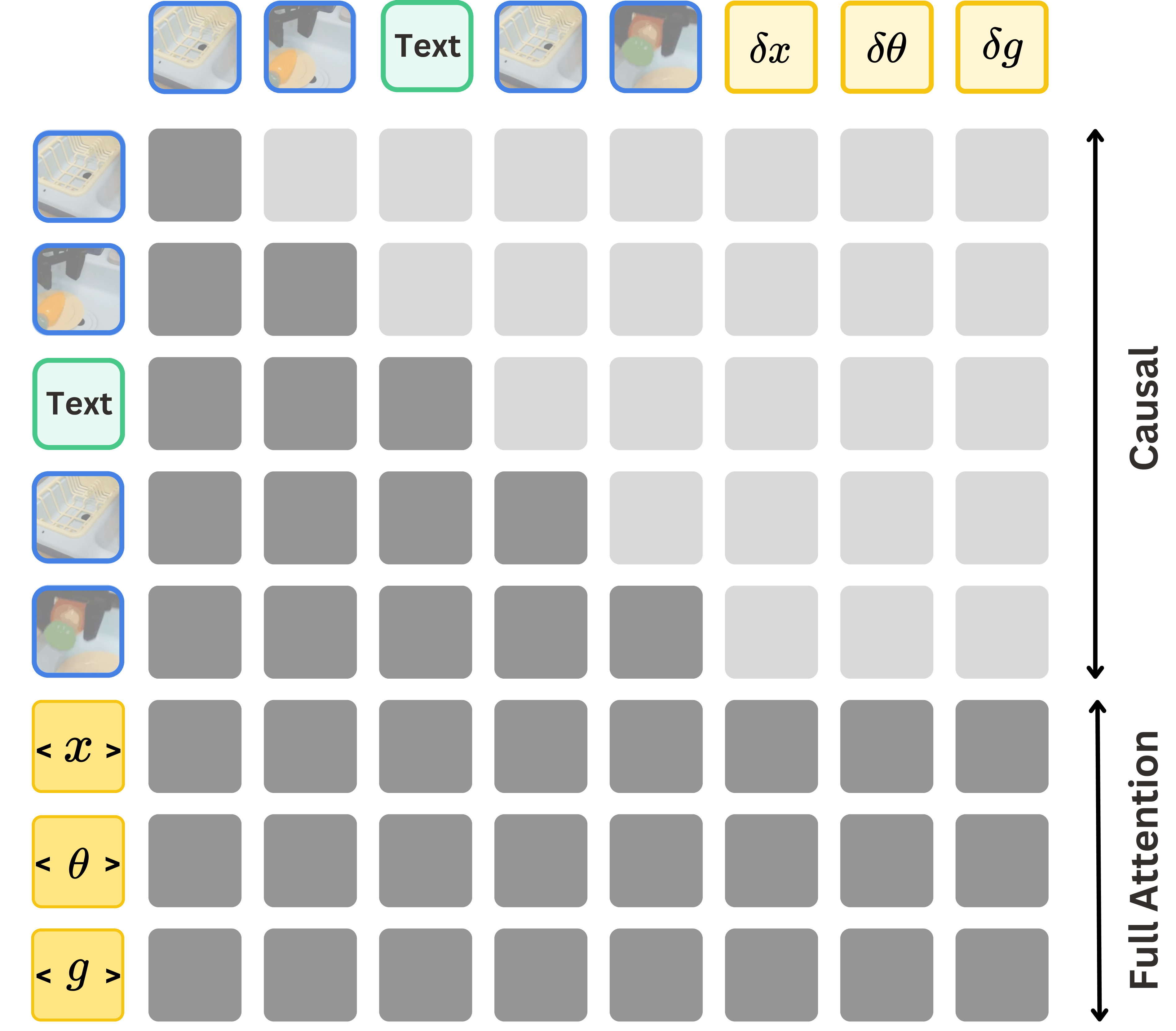}
    \caption{\textbf{Hybrid attention mechanism in \method{}}. We use causal attention for image or text generation and full attention for action generation. $[x]$, $[\theta]$ and $[g]$ are special tokens for parallel decoding of actions.}
    \label{fig:attention}
\end{figure}

\subsection{The Base Vision-Language Model}
\label{sec:vilau}
To enable the visual reasoning capabilities described in Equation ~\eqref{eq:cot}, we build upon VILA-U~\cite{wu2024vila}, an unified multimodal foundation model capable of both understanding and generating image and text tokens.

VILA-U unifies video, image, and language understanding through an autoregressive next-token prediction framework. At its core is a unified vision tower that encodes visual inputs as discrete tokens aligned with textual information. This enables autoregressive image and video generation while significantly enhancing the understanding capabilities of VLMs that leverage discrete visual features. VILA-U utilizes residual quantization~\cite{lee2022autoregressive} to improve the representational capacity of discrete visual features - incorporating a depth transformer, as introduced in RQ-VAE~\cite{lee2022autoregressive}, to gradually predict the residual tokens. The extracted visual features are then passed through a projector before being processed by the LLM backbone. The base model is trained on multimodal pairs including [image, text], [text, image], [video, text], and [text, video]. We use the VILA-U model trained on $256\times 256$ resolution images, where each image is encoded into $16\times 16 \times 4$ tokens with a residual depth of 4~\cite{lee2022autoregressive}. For detailed information about VILA-U training and architecture, we refer readers to~\cite{wu2024vila}.

\subsection{Training Procedures}
\label{sec:trianing}
\begin{table*}[h!]
\centering
\begin{tabular}{lccccc}
\toprule
 & Average ($\uparrow$) & Spatial ($\uparrow$) & Object ($\uparrow$) & Goal ($\uparrow$) & Long ($\uparrow$) \\
\midrule
Diffusion Policy & 72.4 $\pm$ 0.7\% & 78.3 $\pm$ 1.1\% & \textbf{92.5 $\pm$ 0.7\%} & 68.3 $\pm$ 1.2\% & 50.5 $\pm$ 1.3\% \\
Octo fine-tuned & 75.1 $\pm$ 0.6\% & 78.9 $\pm$ 1.0\% & 85.7 $\pm$ 0.9\% & 84.6 $\pm$ 0.9\% & 51.1 $\pm$ 1.3\% \\
OpenVLA fine-tuned & 76.5 $\pm$ 0.6\% & 84.7 $\pm$ 0.9\% & 88.4 $\pm$ 0.8\% & 79.2 $\pm$ 1.0\% & 53.7 $\pm$ 1.3\% \\
CoT-VLA-7B (ours) & \textbf{81.13 $\pm$ 0.6 \%}& \textbf{87.5 $\pm$ 1.4\%} & 91.6 $\pm$ 0.5\% & \textbf{87.6 $\pm$ 0.6\%} & \textbf{69.0 $\pm$ 0.8\%}\\
\bottomrule
\end{tabular}
\caption{\textbf{LIBERO benchmark experimental results.} For each task suite (Spatial, Object, Goal, Long), we report the average success rate and standard error across 3 seeds with 500 episodes each. \method{} achieves the best or competitive performance across all LIBERO benchmarks suites compared to baseline approaches. The bolded entries correspond to highest success rates while underlined entries correspond to second-highest.}
\vspace{-10pt}
\label{tab:libero_performance}
\end{table*}

We pretrain the base 7B VILA-U model on a combination of robot demonstrations $D_r$ and action-less videos $D_v$. During training, we optimize three components, the LLM backbone, projector, and depth transformer, while keeping the vision tower fixed. Our training objective has two key components: subgoal image generation with causal attention \eqref{eq:cot} and action generation with full attention \eqref{eq:cotvla}.

\paragraph{Visual Tokens Prediction}
For subgoal image generation, each training sequence is of form $(l,\mathbf{s}_t,\mathbf{s}_{t+n})$. We follow the training objective used in~\cite{wu2024vila}. At each visual position $j$, the depth transformer, $P_\delta$, autoregressively predicts $D$ residual tokens $(k_{j1}, ..., k_{jD})$ based on the LLM-generated code embedding $h_j$. 
The training objective for visual tokens is then formulated as:
\begin{align}
\mathcal{L}_{\text{visual}} = -\sum_{j} \sum_{d=1}^D \log P_\delta(k_{jd}|k_{j,<d})
\end{align}
where $j$ indexes the positions containing visual tokens. For a more detailed explanation of this loss function, we refer readers to \cite{wu2024vila}, \href{https://arxiv.org/pdf/2409.04429#page=5.5}{Section 3.2}.

\paragraph{Action Tokens Prediction}
For action prediction, each training sequence takes the form $(l,\mathbf{s}_t,\mathbf{s}_{t+n}, \mathbf{a}_{t},...,\mathbf{a}_{t+m})$. Each action $\mathbf{a}_i$ is represented by 7 tokens, with each action dimension independently discretized. Following \cite{kim2024openvla}, we map each continuous action dimension into 256 discrete bins, with bin widths determined by uniformly dividing the interval between the 1st and 99th percentiles of the training data's action distribution. We repurpose the 256 least frequently used tokens in the text tokenizer's vocabulary as action bin tokens. Unlike prior works \cite{driess2023palm,kim2024openvla,o2023open}, we employ full attention for processing and predicting action tokens, enabling all action tokens to interact with each other. This attention mechanism is illustrated in Figure~\ref{fig:attention}. During training, we minimize the cross-entropy loss for action predictions:
\begin{align}
    \mathcal{L}_{\text{action}} = -\sum_{i=1}^{m}\log P_\theta(\mathbf{a}_{t}...\mathbf{a}_{t+m}|l,s_t,s_{t+n})
\end{align}
Given a batch of input sequences, The overall training objective combines the action and visual losses:
\begin{align}
    \mathcal{L} = \mathcal{L}_{\text{action}}+\mathcal{L}_{\text{visual}}
\end{align}

\paragraph{Pretraining Phase}
\label{sec:training_data}
We pretrain \method{} on both robot demonstrations $D_r$ and action-less videos $D_v$ as described in Section~\ref{sec:cot}. For robot demonstrations, we curate a subset of the Open X-Embodiment dataset~\cite{o2023open} (OpenX). Following the pre-processing pipeline established in OpenVLA~\cite{kim2024openvla}, we select and process datasets with third-person camera views and single-arm end-effector control (7-DoF). For action-less videos $D_v$, we incorporate the EPIC-KITCHENS~\cite{kapidis2019egocentric} and Something-Something V2~\cite{goyal2017something} datasets. All images are processed at $256\times 256$ resolution. For visual reasoning, we use subgoal images at future timestep $n$ uniformly sampled from a dataset-specific range $[n_l,n_u]$, where $n_l$ and $n_u$ define the lower and upper bounds of the prediction horizon. We use an action chunk size of 10. For complete dataset specifications and training hyper-parameters, please refer to the supplementary material. 

\paragraph{Adaptation Phase for Downstream Closed-Loop Deployment}
\label{sec:deployment}
For adaptation to downstream tasks, we fine-tune our pretrained model using task-specific robot demonstration data $D_r$ collected on the target robot setups. During this phase, we optimize the LLM backbone, projector, and depth transformer while keeping the vision tower frozen, maintaining the same training setup as the pretraining stage. The resulting model can execute new manipulation tasks based on natural language commands $l$. Algorithm~\ref{alg:test-time} describes our robot control procedure at test time.

\begin{algorithm}[H]
\caption{\method{} test-time closed-loop control}
\begin{algorithmic}
\Require \method{} Model $P_\theta$, initial state $s_0^{\text{obs}}$, language instruction $l$
\State $t \gets 0$
\While{True}
    \State sample $\mathbf{\hat{s}}_{t+n} \sim P_\theta(\mathbf{s}_{t+n} \mid l, \mathbf{s}_t^{\text{obs}})$ 
    \State sample [$\mathbf{\hat{a}}_t,..,\mathbf{\hat{a}}_{t+m}] \sim P_\theta(\mathbf{a}_{t},..,\mathbf{a}_{t+m} \mid l, \mathbf{s}_t^{\text{obs}},\mathbf{s}_{t+n})$ 
    \For{$j = 0$ to $m$}
        \State execute $\mathbf{\hat{a}}_{t+j}$
    \EndFor
    \State $t \gets t + m + 1$
    \State $\mathbf{s}_{t}^{\text{obs}} \gets$ robot observation
\EndWhile
\end{algorithmic}
\label{alg:test-time}
\end{algorithm}

\section{Experiments}
\begin{figure*}[th]
    \centering
    \includegraphics[width=\linewidth]{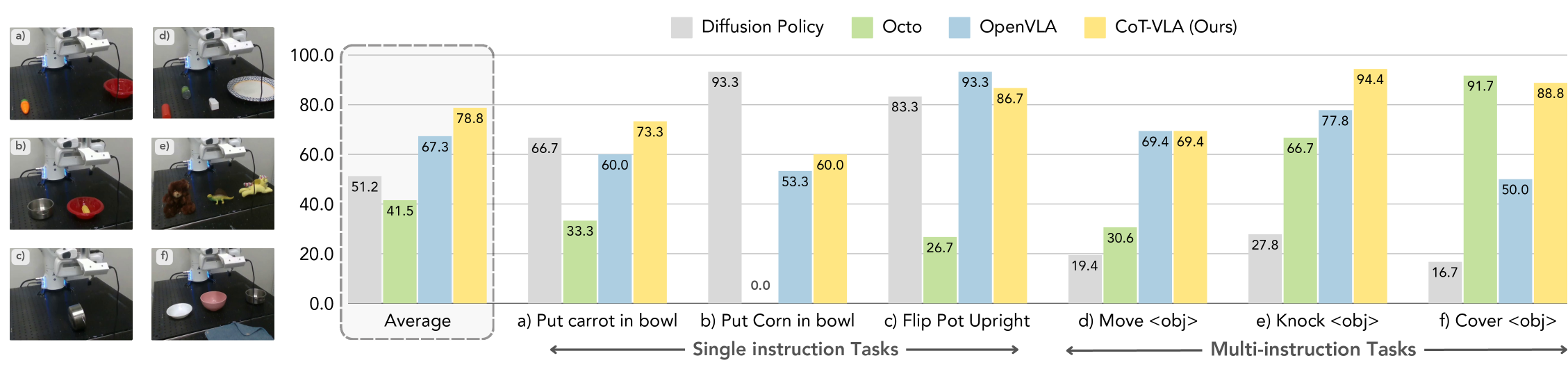}
    \caption{\textbf{Franka-Tabletop comparisons.} Evaluation across six distinct manipulation tasks, with separate models trained per task. Left: Representative initial states for each task setup. Right: Task-specific success rates and cross-task averages for our method and baselines. \method{} achieves best average performance and demonstrates strong capabilities in both single-instruction and multi-instruction scenarios.}
    \label{fig:franka}
    \vspace{-10pt}
\end{figure*}
We evaluate the effectiveness of our approach and our system through a set of experiments spanning both simulation benchmarks and real-world robot manipulation tasks. Our experiments aim to addresses following questions: 
\begin{itemize}
    \item How does our system perform compared to state-of-the-art baselines across multiple benchmarks and embodiments? (Section~\ref{sec:results})
    \item What is the impact of pretraining, visual chain-of-thought reasoning and hybrid attention on task performance? (Section~\ref{sec:ablation})
    \item To what extent does improved generalization in visual reasoning enhance the action prediction capabilities? (Section~\ref{sec:dreamer})
\end{itemize}

\subsection{Experimental Setup}
We conduct evaluations across three complementary settings: the LIBERO benchmark \cite{liu2023libero} for evaluation in simulation environments, the Bridge-V2 platform \cite{walke2023bridgedata} with its dataset of 45k robot demonstrations, and the Franka-Tabletop setup with a stationary, table-mounted Franka Emika Panda robot with limited 10 to 150 robot demonstrations for each testing scenario. 

\paragraph{LIBERO Simulation Benchmark}
We perform evaluation on LIBERO \cite{liu2023libero}, a simulation benchmark comprising four distinct task suites: LIBERO-Spatial, LIBERO-Object, LIBERO-Goal, and LIBERO-Long. Each suite contains 10 diverse tasks with 50 human-teleoperated demonstrations per task, aiming to evaluate the robot's comprehension of spatial relationships, object interactions, and task-specific objectives. We follow the same preprocessed pipeline as in~\cite{kim2024openvla}: (1) removing pause intervals from trajectories, (2) standardizing image resolution to 256×256 pixels, and (3) applying a 180-degree rotation to all images. 

\paragraph{Bridge-V2 Real-Robot Experiments}
We use a 6-DoF WidowX robotic arm, following the experimental setup from Bridge-V2 \cite{walke2023bridgedata}. Our training data has 45k language-annotated trajectories from the Bridge-V2 dataset, encompassing diverse manipulation tasks. While the dataset was incorporated into the pretraining phase alongside OpenX, we performed additional task-specific fine-tuning exclusively on Bridge-V2 until achieving a training action prediction accuracy threshold of 95\%. Following \cite{kim2024openvla}, we evaluate on four tasks designed in~\cite{kim2024openvla} to evaluate visual robustness (varying distractors), motion generalization (novel object positions), semantic generalization (unseen language instruction), and language grounding (instruction following). 

\paragraph{Franka-Tabletop Real-Robot Experiments}
We use a stationary, table-mounted Franka Emika Panda 7-DoF robot arm denoted as Franka-Tabletop. The setup is not seen during the pretraining stage and is designed to assess our model's adaptation capability to novel real-world environments with small amounts of robot demonstrations. We perform evaluations across 6 tasks: 3 narrow domain single-instruction tasks for and 3 diverse multi-instruction tasks outlined in Figure~\ref{fig:franka} and introduced in~\cite{kim2024openvla}.  For each task, the dataset contains between 10 and 150 demonstrations. 

\paragraph{Baselines}
We evaluate our approach against four state-of-the-art baselines. \textbf{Diffusion Policy}~\cite{chi2023diffusion}, a state-of-the-art imitation learning algorithm, is trained from scratch for each test scenario in LIBERO and Franka-Tabletop. The implementation incorporates action chunking and proprioception while conditioning on DistilBERT \cite{sanh2019distilbert} language embeddings. \textbf{OpenVLA}~\cite{kim2024openvla} is an open-source VLA model that fine-tunes pretrained vision-language models on the OpenX dataset; and \textbf{Octo}~\cite{team2024octo} is a generalist model pretrained on OpenX without VLM initialization. For both OpenVLA and Octo, we use their published checkpoints for Bridge-V2 evaluations and fine-tune them for our LIBERO and Franka-Tabletop experiments. \textbf{SUSIE}~\cite{susie}, a two-stage approach, combines instruction-guided image editing for goal generation with a goal-conditioned policy for action generation. We evaluate SUSIE using their published checkpoint on Bridge-V2.

\subsection{Evaluations Results}
\label{sec:results}
\begin{figure*}[t!]
    \centering
    \includegraphics[width=\linewidth]{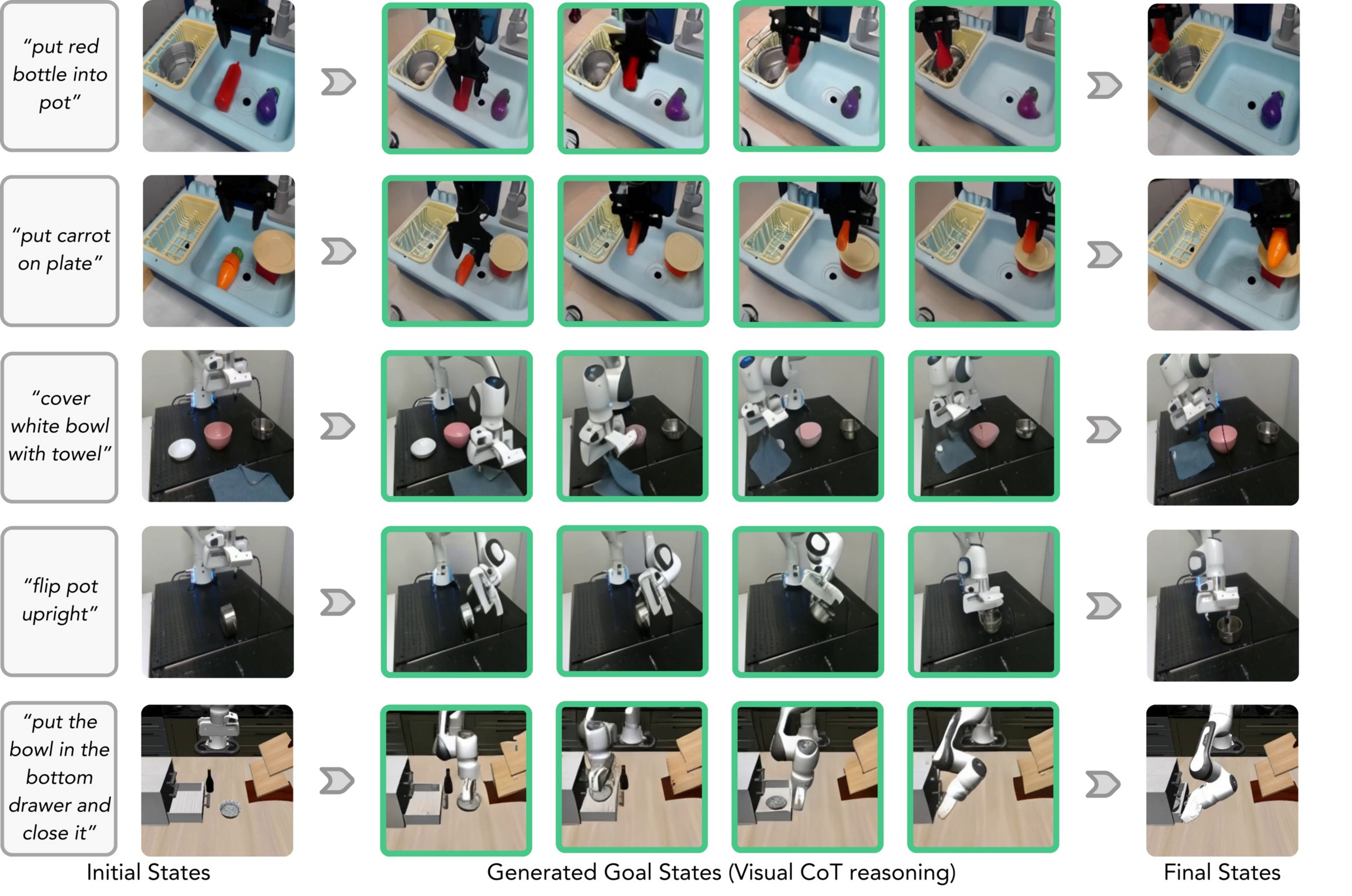}
    \caption{\textbf{Task execution examples for LIBERO, Bridge-V2, and Franka-Tabletop using \method{}.} For each task: Left: text instruction ($l$) and initial state ($s_0^{\text{obs}}$). Middle: generated intermediate goal states ($\hat{s}_t$) demonstrating visual chain-of-thought reasoning, where each goal image is conditioned on both the instruction and the most recent observation. Right: final state ($s_T^{\text{obs}}$) upon task completion. Complete execution trajectories are available in the supplementary video.}
    \label{fig:rollout}
\end{figure*}

\paragraph{LIBERO} We present quantitative results in Table~\ref{tab:libero_performance}, where each method is evaluated over 500 trials per task suite, with 3 random seeds. Success rates are reported with means and standard error.
Qualitative examples of our method's reasoning and execution trajectories are illustrated in Figure~\ref{fig:rollout}. Results demonstrate that \method{} effectively adapts to tasks in the LIBERO simulation environment, achieving best or competitive performance compared to baseline approaches. By analyzing rollout videos of failure cases, we found that baseline methods occasionally overfit to visual cues while disregarding language instructions. Specifically, when initial states appear visually similar across different tasks (e.g., in LIBERO-Spatial), baseline methods execute a different task compared to the commanded task in some episodes. \method{} exhibits better instruction following ability by first reasoning visually about the desired actions via language-grounded subgoal generation, and then predicting the relevant actions for achieving the goal. 

\paragraph{Bridge-V2}
We evaluate \method{} and baselines on the Bridge-V2 benchmark across four generalization categories identified in~\cite{kim2024openvla}: visual generalization (``put eggplant into pot" with cluttered environments), motion generalization (``put carrot on plate" with height variations), semantic generalization (``take purple grapes out of pot"), and language grounding (``put eggplant or red bottle into pot"). We report the quantitative results in Table~\ref{tab:bridge-dataset-comparison}, where each task is tested with 10 trials. SUSIE~\cite{susie} generates visually higher-quality goal images through its diffusion prior (see Section~\ref{sec:limitation} for a detailed discussion on our limitations) but achieves lower success rates on tasks involving novel objects or requiring complex language grounding. Compared to OpenVLA~\cite{kim2024openvla}, \method{} shows slightly lower success rates in visual and language generalization tasks due to grasping failures from action chunking (see Section~\ref{sec:limitation}) rather than errors in visual reasoning. However, \method{} demonstrates competitive performance across all four generalization categories overall, achieving comparable or better results to baseline approaches.

\paragraph{Franka-Tabletop}
We present quantitative results in Table~\ref{fig:franka} and example execution trajectories in Figure~\ref{fig:rollout}. In this experiment, models are fine-tuned on a relatively small set of demonstrations. While Diffusion Policy achieves top performance on single-instruction tasks (e.g., ``put corn in bowl''), its performance degrades on multi-instruction tasks involving diverse objects and complex language instructions. Models pretrained on the OpenX dataset - Octo, OpenVLA, and \method{} - demonstrate better adaptation and performance on multi-instruction tasks where language grounding is critical. Overall, \method{} achieves the highest average performance compared to baseline approaches, showing improvements in both single and multi-instruction scenarios.

\subsection{Ablation Study}
\label{sec:ablation}
\paragraph{Visual CoT, Hybrid Attention, and Action Chunking}
We conduct comprehensive ablation studies on two LIBERO benchmark suites: LIBERO-Spatial and LIBERO-Goal. We evaluate four model variants:
\textbf{VLA} - a baseline implementation following the standard VLA framework \cite{kim2024openvla}, with the same VILA-U backbone but without chain-of-thought reasoning and action chunking; 
\textbf{+ action chunking} - extending the vanilla VLA to predict action sequences of length $m$; 
\textbf{+ hybrid attention} - further adding full attention mechanisms for action sequence prediction, as illustrated in Figure~\ref{fig:attention}; and
\textbf{+ CoT (ours)}: our complete approach with hybrid attention mechanism and visual chain-of-thought reasoning.

\begin{table}
\centering
\begin{tabular}{lcccc}
\toprule
 Category  & SUSIE & Octo & OpenVLA & \method{}\\
\midrule
Visual   & 30\% & 35\%   & \textbf{75\%}   & 65\% \\
Motion   & 10\% & 10\%   & 45\%   & \textbf{60\%} \\
Semantic & 20\% & 0\%    & 40\%   & \textbf{50\%} \\
Language & 40\% & 40\%   & \textbf{75\%}   & 70\% \\
\bottomrule
\end{tabular}
\caption{\textbf{Bridge-V2 Comparison.} Success rates across four generalization categories, with 10 trials per category and partial credit scoring following~\cite{kim2024openvla}. \textbf{Visual}: ``put eggplant into pot" with cluttered environments; \textbf{Motion}: ``put carrot on plate" with height variations; \textbf{Semantic}: ``take purple grapes out of pot"; \textbf{Language}: ``put eggplant or red bottle into pot".}
\label{tab:bridge-dataset-comparison}
\end{table}

\begin{figure}[h]
    \centering
    \vspace{-10pt}
    \includegraphics[width=\linewidth]{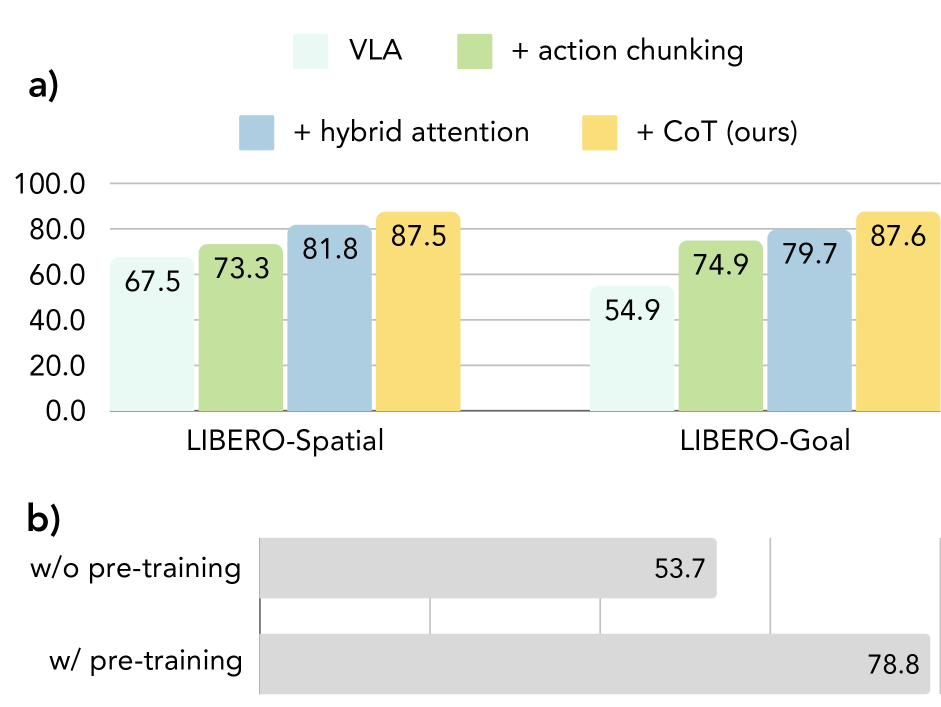}
    \caption{\textbf{Ablation studies of \method{} components.} a) Results on LIBERO-Spatial and LIBERO-Goal benchmarks demonstrate the effectiveness of three components: action chunking, hybrid attention, and visual chain-of-thought reasoning. b) Pretraining ablation experiments on Franka-Tabletop show performance improvements from the OpenX and action-less video pretraining process.}
    \vspace{-10pt}
    \label{fig:ablation}
\end{figure}
As shown in Figure~\ref{fig:ablation}, both benchmark suites demonstrate that action sequence prediction consistently outperforms single-action prediction. The addition of hybrid attention mechanisms further improves performance. Our \method{} achieves the best results validating the effectiveness of visual chain-of-thought reasoning for VLA tasks.

\paragraph{Pretraining}
Our training pipeline has two stages, pretraining VILA-U on the OpenX dataset augmented with action-less video data (Section~\ref{sec:training_data}), and task-specific post-training on robot demonstration data. To assess the importance of our pretraining stage, we conduct ablation studies on the Franka-Tabletop setup. We report the quantitative results in Figure~\ref{fig:ablation}. Our results show that \method{} with our pretraining stage achieves a 46.7\% relative improvement, from 53.7\% to 78.8\%, compared to directly fine-tuning the base VILA-U model on Franka-Tabletop demonstrations, demonstrating better downstream task adaptation.

\subsection{Better Visual Reasoning Helps}
\label{sec:dreamer}
Unlike prior VLAs that only use robot demonstration data $D_{\text{r}}$ during training, \method{} also leverages action-less video data $D_{\text{v}}$ for pretraining through its intermediate visual chain-of-thought reasoning steps. This enables learning of both dynamics and instruction following from captioned videos alone, which are substantially more abundant than robot demonstrations.
To investigate how visual reasoning capabilities transfer to robot performance, we conduct an ablation study on the Franka-Tabletop setup using novel, long-horizon tasks that combine two unseen subtasks. We design two tasks -- (1) ``move the green scallion to the apple-covered book'' and (2) ``move the green cauliflower to the bear-covered book'' -- that are challenging for our model's out-of-distribution generalization. For each task, we collect one demonstration trajectory to obtain ground-truth goal images.
We evaluate each task across 5 trials under two conditions: (1) \method{} using its generated goal images and (2) \method{} using ground-truth goal images from the collected demonstrations.  As shown in Table~\ref{tab:dreamer}, using ground-truth goal images improves the absolute success rate by 40\% for both tasks. This performance boost suggests that advances in visual reasoning and goal image generation could directly translate to better robotic task performance. While our method still struggles with out-of-distribution subgoal generation, recent advances in large-scale video and image models show promising directions for improving visual reasoning capabilities with scaling.

\begin{table}[htbp]
\centering
\begin{tabular}{lccccc}
\toprule
 & Sub-task 1 & Sub-task 2 \\
\midrule
Generated Goal Images & 20\% & 0\% \\
Ground-truth Goal Images & \textbf{60\%} & \textbf{40\%} \\
\bottomrule
\end{tabular}
\caption{\textbf{Better visual reasoning helps.} Success rates comparing \method{} using generated versus ground-truth goal images on out-of-distribution tasks. Results demonstrate that improved visual reasoning (simulated by ground-truth goals) leads to better task performance, suggesting that advances in goal generation can translate to improved action execution.}
\label{tab:dreamer}
\vspace{-10pt}
\end{table}

\section{Conclusion, Limitations and Future Work}
\label{sec:limitation}
In this work, we introduce \method{}, bridging vision-language-action models with chain-of-thought reasoning by introducing intermediate visual goals as explicit reasoning steps. Rather than using abstract representations like bounding boxes or keypoints, we propose using subgoal images sampled from videos as an interpretable and effective intermediate representation. We build our system upon VILA-U, demonstrating strong performance across diverse robotic manipulation tasks.

While our approach demonstrates effectiveness, there are certain limitations. First, generating intermediate image tokens during inference introduces significant computational overhead compared to direct action generation approaches. Our method requires generating 256 image tokens before action tokens, leading to a $7\times$ slowdown on average with an action chunk size of 10. While action chunking and parallel decoding improve inference speed, image generation remains the primary bottleneck. Recent advancement in fast image generation or fast LLM inference techniques could potentially improve the throughput of the model~\cite{chen2024deep,song2023consistency,leviathan2023fast,kwon2023efficient,yu2024an} and be integrated into our system. Second, our autoregressive image generation produces lower visual quality compared to state-of-the-art diffusion-based models. Recent advances in unified multimodal models \cite{xie2024show,Zhou2024TransfusionPT,wu2024janus,wang2024emu3} suggest promising directions for improvements.  Additionally, while effective, our action chunking approach can introduce discontinuous actions between chunks and lacks high-frequency feedback during execution. These limitations could be addressed through temporal smoothing techniques and per-step prediction approaches similar to those proposed in~\cite{chi2023diffusion}. Finally, while \method{} leverages action-less video data during pretraining, current computational constraints limit its ability to achieve visual-reasoning generalization for entirely new tasks. Looking forward, we believe recent advances in video/image generation and world models~\cite{hu2023gaia,kondratyuk2023videopoet,du2024learning,yang2024video,xiang2024pandora} present promising opportunities to enhance generalization capabilities through improved visual reasoning and predictive modeling.

{
    \small
    \bibliographystyle{ieeenat_fullname}
    \bibliography{main}

\begin{thebibliography}{83}
\providecommand{\natexlab}[1]{#1}
\providecommand{\url}[1]{\texttt{#1}}
\expandafter\ifx\csname urlstyle\endcsname\relax
  \providecommand{\doi}[1]{doi: #1}\else
  \providecommand{\doi}{doi: \begingroup \urlstyle{rm}\Url}\fi

\bibitem[Bharadhwaj et~al.(2024)Bharadhwaj, Dwibedi, Gupta, Tulsiani, Doersch, Xiao, Shah, Xia, Sadigh, and Kirmani]{bharadhwaj2024gen2act}
Homanga Bharadhwaj, Debidatta Dwibedi, Abhinav Gupta, Shubham Tulsiani, Carl Doersch, Ted Xiao, Dhruv Shah, Fei Xia, Dorsa Sadigh, and Sean Kirmani.
\newblock Gen2act: Human video generation in novel scenarios enables generalizable robot manipulation.
\newblock \emph{arXiv preprint arXiv:2409.16283}, 2024.

\bibitem[Black et~al.(2023)Black, Nakamoto, Atreya, Walke, Finn, Kumar, and Levine]{susie}
Kevin Black, Mitsuhiko Nakamoto, Pranav Atreya, Homer Walke, Chelsea Finn, Aviral Kumar, and Sergey Levine.
\newblock Zero-shot robotic manipulation with pretrained image-editing diffusion models.
\newblock \emph{arXiv preprint arXiv:2310.10639}, 2023.

\bibitem[Brohan et~al.(2022)Brohan, Brown, Carbajal, Chebotar, Dabis, Finn, Gopalakrishnan, Hausman, Herzog, Hsu, et~al.]{brohan2022rt}
Anthony Brohan, Noah Brown, Justice Carbajal, Yevgen Chebotar, Joseph Dabis, Chelsea Finn, Keerthana Gopalakrishnan, Karol Hausman, Alex Herzog, Jasmine Hsu, et~al.
\newblock Rt-1: Robotics transformer for real-world control at scale.
\newblock \emph{arXiv preprint arXiv:2212.06817}, 2022.

\bibitem[Caron et~al.(2021)Caron, Touvron, Misra, J{\'e}gou, Mairal, Bojanowski, and Joulin]{caron2021emerging}
Mathilde Caron, Hugo Touvron, Ishan Misra, Herv{\'e} J{\'e}gou, Julien Mairal, Piotr Bojanowski, and Armand Joulin.
\newblock Emerging properties in self-supervised vision transformers.
\newblock In \emph{Proceedings of the IEEE/CVF international conference on computer vision}, pages 9650--9660, 2021.

\bibitem[Cheang et~al.(2024)Cheang, Chen, Jing, Kong, Li, Li, Liu, Wu, Xu, Yang, et~al.]{cheang2024gr}
Chi-Lam Cheang, Guangzeng Chen, Ya Jing, Tao Kong, Hang Li, Yifeng Li, Yuxiao Liu, Hongtao Wu, Jiafeng Xu, Yichu Yang, et~al.
\newblock Gr-2: A generative video-language-action model with web-scale knowledge for robot manipulation.
\newblock \emph{arXiv preprint arXiv:2410.06158}, 2024.

\bibitem[Chen et~al.(2024{\natexlab{a}})Chen, Xu, Kirmani, Ichter, Sadigh, Guibas, and Xia]{Chen_2024_CVPR}
Boyuan Chen, Zhuo Xu, Sean Kirmani, Brain Ichter, Dorsa Sadigh, Leonidas Guibas, and Fei Xia.
\newblock Spatialvlm: Endowing vision-language models with spatial reasoning capabilities.
\newblock In \emph{Proceedings of the IEEE/CVF Conference on Computer Vision and Pattern Recognition (CVPR)}, pages 14455--14465, 2024{\natexlab{a}}.

\bibitem[Chen et~al.(2024{\natexlab{b}})Chen, Cai, Chen, Xie, Yang, Tang, Li, Lu, and Han]{chen2024deep}
Junyu Chen, Han Cai, Junsong Chen, Enze Xie, Shang Yang, Haotian Tang, Muyang Li, Yao Lu, and Song Han.
\newblock Deep compression autoencoder for efficient high-resolution diffusion models.
\newblock \emph{arXiv preprint arXiv:2410.10733}, 2024{\natexlab{b}}.

\bibitem[Chen et~al.()Chen, Adebola, and Goldberg]{BerkeleyUR5Website}
Lawrence~Yunliang Chen, Simeon Adebola, and Ken Goldberg.
\newblock Berkeley {UR5} demonstration dataset.
\newblock \url{https://sites.google.com/view/berkeley-ur5/home}.

\bibitem[Chen et~al.(2023)Chen, Wang, Beyer, Kolesnikov, Wu, Voigtlaender, Mustafa, Goodman, Alabdulmohsin, Padlewski, et~al.]{chen2023pali}
Xi Chen, Xiao Wang, Lucas Beyer, Alexander Kolesnikov, Jialin Wu, Paul Voigtlaender, Basil Mustafa, Sebastian Goodman, Ibrahim Alabdulmohsin, Piotr Padlewski, et~al.
\newblock Pali-3 vision language models: Smaller, faster, stronger.
\newblock \emph{arXiv preprint arXiv:2310.09199}, 2023.

\bibitem[Chi et~al.(2023)Chi, Xu, Feng, Cousineau, Du, Burchfiel, Tedrake, and Song]{chi2023diffusion}
Cheng Chi, Zhenjia Xu, Siyuan Feng, Eric Cousineau, Yilun Du, Benjamin Burchfiel, Russ Tedrake, and Shuran Song.
\newblock Diffusion policy: Visuomotor policy learning via action diffusion.
\newblock \emph{The International Journal of Robotics Research}, page 02783649241273668, 2023.

\bibitem[Ding et~al.(2019)Ding, Florensa, Abbeel, and Phielipp]{ding2019goal}
Yiming Ding, Carlos Florensa, Pieter Abbeel, and Mariano Phielipp.
\newblock Goal-conditioned imitation learning.
\newblock \emph{Advances in neural information processing systems}, 32, 2019.

\bibitem[Driess et~al.(2023)Driess, Xia, Sajjadi, Lynch, Chowdhery, Ichter, Wahid, Tompson, Vuong, Yu, et~al.]{driess2023palm}
Danny Driess, Fei Xia, Mehdi~SM Sajjadi, Corey Lynch, Aakanksha Chowdhery, Brian Ichter, Ayzaan Wahid, Jonathan Tompson, Quan Vuong, Tianhe Yu, et~al.
\newblock Palm-e: An embodied multimodal language model.
\newblock \emph{arXiv preprint arXiv:2303.03378}, 2023.

\bibitem[Du et~al.(2023{\natexlab{a}})Du, Konyushkova, Denil, Raju, Landon, Hill, de~Freitas, and Cabi]{du2023reward}
Yuqing Du, Ksenia Konyushkova, Misha Denil, Akhil Raju, Jessica Landon, Felix Hill, Nando de Freitas, and Serkan Cabi.
\newblock Vision-language models as success detectors.
\newblock \emph{arXiv preprint arXiv:2303.07280}, 2023{\natexlab{a}}.

\bibitem[Du et~al.(2023{\natexlab{b}})Du, Yang, Florence, Xia, Wahid, Ichter, Sermanet, Yu, Abbeel, Tenenbaum, et~al.]{du2023video}
Yilun Du, Mengjiao Yang, Pete Florence, Fei Xia, Ayzaan Wahid, Brian Ichter, Pierre Sermanet, Tianhe Yu, Pieter Abbeel, Joshua~B Tenenbaum, et~al.
\newblock Video language planning.
\newblock \emph{arXiv preprint arXiv:2310.10625}, 2023{\natexlab{b}}.

\bibitem[Du et~al.(2024)Du, Yang, Dai, Dai, Nachum, Tenenbaum, Schuurmans, and Abbeel]{du2024learning}
Yilun Du, Sherry Yang, Bo Dai, Hanjun Dai, Ofir Nachum, Josh Tenenbaum, Dale Schuurmans, and Pieter Abbeel.
\newblock Learning universal policies via text-guided video generation.
\newblock \emph{Advances in Neural Information Processing Systems}, 36, 2024.

\bibitem[Ebert et~al.(2021)Ebert, Yang, Schmeckpeper, Bucher, Georgakis, Daniilidis, Finn, and Levine]{ebert2021bridge}
Frederik Ebert, Yanlai Yang, Karl Schmeckpeper, Bernadette Bucher, Georgios Georgakis, Kostas Daniilidis, Chelsea Finn, and Sergey Levine.
\newblock Bridge data: Boosting generalization of robotic skills with cross-domain datasets.
\newblock \emph{arXiv preprint arXiv:2109.13396}, 2021.

\bibitem[Fu et~al.(2024{\natexlab{a}})Fu, Zhao, Wu, Wetzstein, and Finn]{fu2024humanplus}
Zipeng Fu, Qingqing Zhao, Qi Wu, Gordon Wetzstein, and Chelsea Finn.
\newblock Humanplus: Humanoid shadowing and imitation from humans.
\newblock \emph{arXiv preprint arXiv:2406.10454}, 2024{\natexlab{a}}.

\bibitem[Fu et~al.(2024{\natexlab{b}})Fu, Zhao, and Finn]{fu2024mobile}
Zipeng Fu, Tony~Z. Zhao, and Chelsea Finn.
\newblock Mobile aloha: Learning bimanual mobile manipulation with low-cost whole-body teleoperation.
\newblock In \emph{{Conference on Robot Learning (CoRL)}}, 2024{\natexlab{b}}.

\bibitem[Gadre et~al.(2023)Gadre, Wortsman, Ilharco, Schmidt, and Song]{gadre2023cows}
Samir~Yitzhak Gadre, Mitchell Wortsman, Gabriel Ilharco, Ludwig Schmidt, and Shuran Song.
\newblock Cows on pasture: Baselines and benchmarks for language-driven zero-shot object navigation.
\newblock In \emph{Proceedings of the IEEE/CVF Conference on Computer Vision and Pattern Recognition}, pages 23171--23181, 2023.

\bibitem[Goyal et~al.(2017)Goyal, Ebrahimi~Kahou, Michalski, Materzynska, Westphal, Kim, Haenel, Fruend, Yianilos, Mueller-Freitag, et~al.]{goyal2017something}
Raghav Goyal, Samira Ebrahimi~Kahou, Vincent Michalski, Joanna Materzynska, Susanne Westphal, Heuna Kim, Valentin Haenel, Ingo Fruend, Peter Yianilos, Moritz Mueller-Freitag, et~al.
\newblock The" something something" video database for learning and evaluating visual common sense.
\newblock In \emph{Proceedings of the IEEE international conference on computer vision}, pages 5842--5850, 2017.

\bibitem[Ha et~al.(2023)Ha, Florence, and Song]{ha2023scaling}
Huy Ha, Pete Florence, and Shuran Song.
\newblock Scaling up and distilling down: Language-guided robot skill acquisition.
\newblock In \emph{Conference on Robot Learning}, pages 3766--3777. PMLR, 2023.

\bibitem[Harvey and Wood(2023)]{harvey2023visual}
William Harvey and Frank Wood.
\newblock Visual chain-of-thought diffusion models.
\newblock \emph{arXiv preprint arXiv:2303.16187}, 2023.

\bibitem[Hu et~al.(2023)Hu, Russell, Yeo, Murez, Fedoseev, Kendall, Shotton, and Corrado]{hu2023gaia}
Anthony Hu, Lloyd Russell, Hudson Yeo, Zak Murez, George Fedoseev, Alex Kendall, Jamie Shotton, and Gianluca Corrado.
\newblock Gaia-1: A generative world model for autonomous driving.
\newblock \emph{arXiv preprint arXiv:2309.17080}, 2023.

\bibitem[Hu et~al.(2024)Hu, Shi, Fu, Roth, Ostendorf, Zettlemoyer, Smith, and Krishna]{hu2024visual}
Yushi Hu, Weijia Shi, Xingyu Fu, Dan Roth, Mari Ostendorf, Luke Zettlemoyer, Noah~A Smith, and Ranjay Krishna.
\newblock Visual sketchpad: Sketching as a visual chain of thought for multimodal language models.
\newblock \emph{arXiv preprint arXiv:2406.09403}, 2024.

\bibitem[Huang et~al.(2023)Huang, Mees, Zeng, and Burgard]{huang2023visual}
Chenguang Huang, Oier Mees, Andy Zeng, and Wolfram Burgard.
\newblock Visual language maps for robot navigation.
\newblock In \emph{2023 IEEE International Conference on Robotics and Automation (ICRA)}, pages 10608--10615. IEEE, 2023.

\bibitem[Huang et~al.(2024)Huang, Wang, Li, Zhang, and Fei-Fei]{huang2024rekep}
Wenlong Huang, Chen Wang, Yunzhu Li, Ruohan Zhang, and Li Fei-Fei.
\newblock Rekep: Spatio-temporal reasoning of relational keypoint constraints for robotic manipulation.
\newblock \emph{arXiv preprint arXiv:2409.01652}, 2024.

\bibitem[Kapidis et~al.(2019)Kapidis, Poppe, Van~Dam, Noldus, and Veltkamp]{kapidis2019egocentric}
Georgios Kapidis, Ronald Poppe, Elsbeth Van~Dam, Lucas Noldus, and Remco Veltkamp.
\newblock Egocentric hand track and object-based human action recognition.
\newblock In \emph{2019 IEEE SmartWorld, Ubiquitous Intelligence \& Computing, Advanced \& Trusted Computing, Scalable Computing \& Communications, Cloud \& Big Data Computing, Internet of People and Smart City Innovation (SmartWorld/SCALCOM/UIC/ATC/CBDCom/IOP/SCI)}, pages 922--929. IEEE, 2019.

\bibitem[Karamcheti et~al.(2024)Karamcheti, Nair, Balakrishna, Liang, Kollar, and Sadigh]{karamcheti2024prismatic}
Siddharth Karamcheti, Suraj Nair, Ashwin Balakrishna, Percy Liang, Thomas Kollar, and Dorsa Sadigh.
\newblock Prismatic vlms: Investigating the design space of visually-conditioned language models.
\newblock \emph{arXiv preprint arXiv:2402.07865}, 2024.

\bibitem[Kim et~al.(2024)Kim, Pertsch, Karamcheti, Xiao, Balakrishna, Nair, Rafailov, Foster, Lam, Sanketi, et~al.]{kim2024openvla}
Moo~Jin Kim, Karl Pertsch, Siddharth Karamcheti, Ted Xiao, Ashwin Balakrishna, Suraj Nair, Rafael Rafailov, Ethan Foster, Grace Lam, Pannag Sanketi, et~al.
\newblock Openvla: An open-source vision-language-action model.
\newblock \emph{arXiv preprint arXiv:2406.09246}, 2024.

\bibitem[Kondratyuk et~al.(2023)Kondratyuk, Yu, Gu, Lezama, Huang, Schindler, Hornung, Birodkar, Yan, Chiu, et~al.]{kondratyuk2023videopoet}
Dan Kondratyuk, Lijun Yu, Xiuye Gu, Jos{\'e} Lezama, Jonathan Huang, Grant Schindler, Rachel Hornung, Vighnesh Birodkar, Jimmy Yan, Ming-Chang Chiu, et~al.
\newblock Videopoet: A large language model for zero-shot video generation.
\newblock \emph{arXiv preprint arXiv:2312.14125}, 2023.

\bibitem[Kwon et~al.(2023)Kwon, Li, Zhuang, Sheng, Zheng, Yu, Gonzalez, Zhang, and Stoica]{kwon2023efficient}
Woosuk Kwon, Zhuohan Li, Siyuan Zhuang, Ying Sheng, Lianmin Zheng, Cody~Hao Yu, Joseph~E. Gonzalez, Hao Zhang, and Ion Stoica.
\newblock Efficient memory management for large language model serving with pagedattention.
\newblock In \emph{Proceedings of the ACM SIGOPS 29th Symposium on Operating Systems Principles}, 2023.

\bibitem[Lee et~al.(2022)Lee, Kim, Kim, Cho, and Han]{lee2022autoregressive}
Doyup Lee, Chiheon Kim, Saehoon Kim, Minsu Cho, and Wook-Shin Han.
\newblock Autoregressive image generation using residual quantization.
\newblock In \emph{Proceedings of the IEEE/CVF Conference on Computer Vision and Pattern Recognition}, pages 11523--11532, 2022.

\bibitem[Leviathan et~al.(2023)Leviathan, Kalman, and Matias]{leviathan2023fast}
Yaniv Leviathan, Matan Kalman, and Yossi Matias.
\newblock Fast inference from transformers via speculative decoding.
\newblock In \emph{International Conference on Machine Learning}, pages 19274--19286. PMLR, 2023.

\bibitem[Li et~al.(2024)Li, Wang, Mao, Ivanovic, Veer, Leung, and Pavone]{li2024driving}
Boyi Li, Yue Wang, Jiageng Mao, Boris Ivanovic, Sushant Veer, Karen Leung, and Marco Pavone.
\newblock Driving everywhere with large language model policy adaptation.
\newblock In \emph{Proceedings of the IEEE/CVF Conference on Computer Vision and Pattern Recognition}, pages 14948--14957, 2024.

\bibitem[Liang et~al.(2024)Liang, Liu, Ozguroglu, Sudhakar, Dave, Tokmakov, Song, and Vondrick]{liang2024dreamitate}
Junbang Liang, Ruoshi Liu, Ege Ozguroglu, Sruthi Sudhakar, Achal Dave, Pavel Tokmakov, Shuran Song, and Carl Vondrick.
\newblock Dreamitate: Real-world visuomotor policy learning via video generation.
\newblock \emph{arXiv preprint arXiv:2406.16862}, 2024.

\bibitem[Lin et~al.(2024)Lin, Hu, Sheng, Wen, You, and Gao]{lin2024data}
Fanqi Lin, Yingdong Hu, Pingyue Sheng, Chuan Wen, Jiacheng You, and Yang Gao.
\newblock Data scaling laws in imitation learning for robotic manipulation.
\newblock \emph{arXiv preprint arXiv:2410.18647}, 2024.

\bibitem[Liu et~al.(2023)Liu, Zhu, Gao, Feng, Liu, Zhu, and Stone]{liu2023libero}
Bo Liu, Yifeng Zhu, Chongkai Gao, Yihao Feng, Qiang Liu, Yuke Zhu, and Peter Stone.
\newblock Libero: Benchmarking knowledge transfer for lifelong robot learning.
\newblock \emph{arXiv preprint arXiv:2306.03310}, 2023.

\bibitem[Liu et~al.(2024)Liu, Li, Wu, and Lee]{liu2024visual}
Haotian Liu, Chunyuan Li, Qingyang Wu, and Yong~Jae Lee.
\newblock Visual instruction tuning.
\newblock \emph{Advances in neural information processing systems}, 36, 2024.

\bibitem[Lu et~al.(2024)Lu, Clark, Lee, Zhang, Khosla, Marten, Hoiem, and Kembhavi]{lu2024unified}
Jiasen Lu, Christopher Clark, Sangho Lee, Zichen Zhang, Savya Khosla, Ryan Marten, Derek Hoiem, and Aniruddha Kembhavi.
\newblock Unified-io 2: Scaling autoregressive multimodal models with vision language audio and action.
\newblock In \emph{Proceedings of the IEEE/CVF Conference on Computer Vision and Pattern Recognition}, pages 26439--26455, 2024.

\bibitem[Ma et~al.(2023)Ma, Liang, Wang, Huang, Bastani, Jayaraman, Zhu, Fan, and Anandkumar]{ma2023eureka}
Yecheng~Jason Ma, William Liang, Guanzhi Wang, De-An Huang, Osbert Bastani, Dinesh Jayaraman, Yuke Zhu, Linxi Fan, and Anima Anandkumar.
\newblock Eureka: Human-level reward design via coding large language models.
\newblock \emph{arXiv preprint arXiv: Arxiv-2310.12931}, 2023.

\bibitem[Majumdar et~al.(2023)Majumdar, Yadav, Arnaud, Ma, Chen, Silwal, Jain, Berges, Wu, Vakil, et~al.]{majumdar2023we}
Arjun Majumdar, Karmesh Yadav, Sergio Arnaud, Jason Ma, Claire Chen, Sneha Silwal, Aryan Jain, Vincent-Pierre Berges, Tingfan Wu, Jay Vakil, et~al.
\newblock Where are we in the search for an artificial visual cortex for embodied intelligence?
\newblock \emph{Advances in Neural Information Processing Systems}, 36:\penalty0 655--677, 2023.

\bibitem[Mandlekar et~al.(2018)Mandlekar, Zhu, Garg, Booher, Spero, Tung, Gao, Emmons, Gupta, Orbay, et~al.]{mandlekar2018roboturk}
Ajay Mandlekar, Yuke Zhu, Animesh Garg, Jonathan Booher, Max Spero, Albert Tung, Julian Gao, John Emmons, Anchit Gupta, Emre Orbay, et~al.
\newblock Roboturk: A crowdsourcing platform for robotic skill learning through imitation.
\newblock In \emph{Conference on Robot Learning}, pages 879--893. PMLR, 2018.

\bibitem[Mees et~al.(2023)Mees, Borja-Diaz, and Burgard]{taco2}
Oier Mees, Jessica Borja-Diaz, and Wolfram Burgard.
\newblock Grounding language with visual affordances over unstructured data.
\newblock In \emph{2023 IEEE International Conference on Robotics and Automation (ICRA)}, pages 11576--11582. IEEE, 2023.

\bibitem[Micha{\l} et~al.(2024)Micha{\l}, William, Karl, Oier, Chelsea, and Sergey]{michal2024robotic}
Zawalski Micha{\l}, Chen William, Pertsch Karl, Mees Oier, Finn Chelsea, and Levine Sergey.
\newblock Robotic control via embodied chain-of-thought reasoning.
\newblock \emph{arXiv preprint arXiv:2407.08693}, 2024.

\bibitem[Mu et~al.(2024)Mu, Zhang, Hu, Wang, Ding, Jin, Wang, Dai, Qiao, and Luo]{mu2024embodiedgpt}
Yao Mu, Qinglong Zhang, Mengkang Hu, Wenhai Wang, Mingyu Ding, Jun Jin, Bin Wang, Jifeng Dai, Yu Qiao, and Ping Luo.
\newblock Embodiedgpt: Vision-language pre-training via embodied chain of thought.
\newblock \emph{Advances in Neural Information Processing Systems}, 36, 2024.

\bibitem[Nair et~al.(2018)Nair, Pong, Dalal, Bahl, Lin, and Levine]{nair2018visual}
Ashvin~V Nair, Vitchyr Pong, Murtaza Dalal, Shikhar Bahl, Steven Lin, and Sergey Levine.
\newblock Visual reinforcement learning with imagined goals.
\newblock \emph{Advances in neural information processing systems}, 31, 2018.

\bibitem[Ni et~al.(2024)Ni, Hao, Wu, Kou, Liu, Zheng, Wang, and Zhuang]{ni2024generate}
Fei Ni, Jianye Hao, Shiguang Wu, Longxin Kou, Jiashun Liu, Yan Zheng, Bin Wang, and Yuzheng Zhuang.
\newblock Generate subgoal images before act: Unlocking the chain-of-thought reasoning in diffusion model for robot manipulation with multimodal prompts.
\newblock In \emph{Proceedings of the IEEE/CVF Conference on Computer Vision and Pattern Recognition}, 2024.

\bibitem[O'Neill et~al.(2023)O'Neill, Rehman, Gupta, Maddukuri, Gupta, Padalkar, Lee, Pooley, Gupta, Mandlekar, et~al.]{o2023open}
Abby O'Neill, Abdul Rehman, Abhinav Gupta, Abhiram Maddukuri, Abhishek Gupta, Abhishek Padalkar, Abraham Lee, Acorn Pooley, Agrim Gupta, Ajay Mandlekar, et~al.
\newblock Open x-embodiment: Robotic learning datasets and rt-x models.
\newblock \emph{arXiv preprint arXiv:2310.08864}, 2023.

\bibitem[Radford et~al.(2021)Radford, Kim, Hallacy, Ramesh, Goh, Agarwal, Sastry, Askell, Mishkin, Clark, et~al.]{radford2021learning}
Alec Radford, Jong~Wook Kim, Chris Hallacy, Aditya Ramesh, Gabriel Goh, Sandhini Agarwal, Girish Sastry, Amanda Askell, Pamela Mishkin, Jack Clark, et~al.
\newblock Learning transferable visual models from natural language supervision.
\newblock In \emph{International conference on machine learning}, pages 8748--8763. PMLR, 2021.

\bibitem[Rose et~al.(2023)Rose, Himakunthala, Ouyang, He, Mei, Lu, Saxon, Sonar, Mirza, and Wang]{rose2023visual}
Daniel Rose, Vaishnavi Himakunthala, Andy Ouyang, Ryan He, Alex Mei, Yujie Lu, Michael Saxon, Chinmay Sonar, Diba Mirza, and William~Yang Wang.
\newblock Visual chain of thought: bridging logical gaps with multimodal infillings.
\newblock \emph{arXiv preprint arXiv:2305.02317}, 2023.

\bibitem[Rosete-Beas et~al.(2023)Rosete-Beas, Mees, Kalweit, Boedecker, and Burgard]{taco1}
Erick Rosete-Beas, Oier Mees, Gabriel Kalweit, Joschka Boedecker, and Wolfram Burgard.
\newblock Latent plans for task-agnostic offline reinforcement learning.
\newblock In \emph{Conference on Robot Learning}, pages 1838--1849. PMLR, 2023.

\bibitem[Sanh(2019)]{sanh2019distilbert}
V Sanh.
\newblock Distilbert, a distilled version of bert: smaller, faster, cheaper and lighter.
\newblock \emph{arXiv preprint arXiv:1910.01108}, 2019.

\bibitem[Shao et~al.(2024)Shao, Qian, Xiao, Song, Zong, Wang, Liu, and Li]{shao2024visual}
Hao Shao, Shengju Qian, Han Xiao, Guanglu Song, Zhuofan Zong, Letian Wang, Yu Liu, and Hongsheng Li.
\newblock Visual cot: Unleashing chain-of-thought reasoning in multi-modal language models.
\newblock \emph{arXiv preprint arXiv:2403.16999}, 2024.

\bibitem[Shridhar et~al.(2022)Shridhar, Manuelli, and Fox]{shridhar2022cliport}
Mohit Shridhar, Lucas Manuelli, and Dieter Fox.
\newblock Cliport: What and where pathways for robotic manipulation.
\newblock In \emph{Conference on robot learning}, pages 894--906. PMLR, 2022.

\bibitem[Shridhar et~al.(2024)Shridhar, Lo, and James]{shridhar2024generative}
Mohit Shridhar, Yat~Long Lo, and Stephen James.
\newblock Generative image as action models.
\newblock \emph{arXiv preprint arXiv:2407.07875}, 2024.

\bibitem[Singh et~al.(2023)Singh, Blukis, Mousavian, Goyal, Xu, Tremblay, Fox, Thomason, and Garg]{singh2023progprompt}
Ishika Singh, Valts Blukis, Arsalan Mousavian, Ankit Goyal, Danfei Xu, Jonathan Tremblay, Dieter Fox, Jesse Thomason, and Animesh Garg.
\newblock Progprompt: Generating situated robot task plans using large language models.
\newblock In \emph{2023 IEEE International Conference on Robotics and Automation (ICRA)}, pages 11523--11530. IEEE, 2023.

\bibitem[Song et~al.(2023)Song, Dhariwal, Chen, and Sutskever]{song2023consistency}
Yang Song, Prafulla Dhariwal, Mark Chen, and Ilya Sutskever.
\newblock Consistency models.
\newblock \emph{arXiv preprint arXiv:2303.01469}, 2023.

\bibitem[Team(2024)]{team2024chameleon}
Chameleon Team.
\newblock Chameleon: Mixed-modal early-fusion foundation models.
\newblock \emph{arXiv preprint arXiv:2405.09818}, 2024.

\bibitem[Team et~al.(2024)Team, Ghosh, Walke, Pertsch, Black, Mees, Dasari, Hejna, Kreiman, Xu, et~al.]{team2024octo}
Octo~Model Team, Dibya Ghosh, Homer Walke, Karl Pertsch, Kevin Black, Oier Mees, Sudeep Dasari, Joey Hejna, Tobias Kreiman, Charles Xu, et~al.
\newblock Octo: An open-source generalist robot policy.
\newblock \emph{arXiv preprint arXiv:2405.12213}, 2024.

\bibitem[Walke et~al.(2023)Walke, Black, Zhao, Vuong, Zheng, Hansen-Estruch, He, Myers, Kim, Du, et~al.]{walke2023bridgedata}
Homer~Rich Walke, Kevin Black, Tony~Z Zhao, Quan Vuong, Chongyi Zheng, Philippe Hansen-Estruch, Andre~Wang He, Vivek Myers, Moo~Jin Kim, Max Du, et~al.
\newblock Bridgedata v2: A dataset for robot learning at scale.
\newblock In \emph{Conference on Robot Learning}, pages 1723--1736. PMLR, 2023.

\bibitem[Wang et~al.(2024)Wang, Zhang, Luo, Sun, Cui, Wang, Zhang, Wang, Li, Yu, et~al.]{wang2024emu3}
Xinlong Wang, Xiaosong Zhang, Zhengxiong Luo, Quan Sun, Yufeng Cui, Jinsheng Wang, Fan Zhang, Yueze Wang, Zhen Li, Qiying Yu, et~al.
\newblock Emu3: Next-token prediction is all you need.
\newblock \emph{arXiv preprint arXiv:2409.18869}, 2024.

\bibitem[Wei et~al.(2022)Wei, Wang, Schuurmans, Bosma, Xia, Chi, Le, Zhou, et~al.]{wei2022chain}
Jason Wei, Xuezhi Wang, Dale Schuurmans, Maarten Bosma, Fei Xia, Ed Chi, Quoc~V Le, Denny Zhou, et~al.
\newblock Chain-of-thought prompting elicits reasoning in large language models.
\newblock \emph{Advances in neural information processing systems}, 2022.

\bibitem[Wen et~al.(2023)Wen, Lin, So, Chen, Dou, Gao, and Abbeel]{wen2023any}
Chuan Wen, Xingyu Lin, John So, Kai Chen, Qi Dou, Yang Gao, and Pieter Abbeel.
\newblock Any-point trajectory modeling for policy learning.
\newblock \emph{arXiv preprint arXiv:2401.00025}, 2023.

\bibitem[Wen et~al.(2024)Wen, Zhu, Li, Zhu, Wu, Xu, Cheng, Shen, Peng, Feng, et~al.]{wen2024tinyvla}
Junjie Wen, Yichen Zhu, Jinming Li, Minjie Zhu, Kun Wu, Zhiyuan Xu, Ran Cheng, Chaomin Shen, Yaxin Peng, Feifei Feng, et~al.
\newblock Tinyvla: Towards fast, data-efficient vision-language-action models for robotic manipulation.
\newblock \emph{arXiv preprint arXiv:2409.12514}, 2024.

\bibitem[Wu et~al.(2024{\natexlab{a}})Wu, Chen, Wu, Ma, Liu, Pan, Liu, Xie, Yu, Ruan, et~al.]{wu2024janus}
Chengyue Wu, Xiaokang Chen, Zhiyu Wu, Yiyang Ma, Xingchao Liu, Zizheng Pan, Wen Liu, Zhenda Xie, Xingkai Yu, Chong Ruan, et~al.
\newblock Janus: Decoupling visual encoding for unified multimodal understanding and generation.
\newblock \emph{arXiv preprint arXiv:2410.13848}, 2024{\natexlab{a}}.

\bibitem[Wu et~al.(2023)Wu, Jing, Cheang, Chen, Xu, Li, Liu, Li, and Kong]{wu2023unleashing}
Hongtao Wu, Ya Jing, Chilam Cheang, Guangzeng Chen, Jiafeng Xu, Xinghang Li, Minghuan Liu, Hang Li, and Tao Kong.
\newblock Unleashing large-scale video generative pre-training for visual robot manipulation.
\newblock \emph{arXiv preprint arXiv:2312.13139}, 2023.

\bibitem[Wu et~al.(2024{\natexlab{b}})Wu, Zhang, Chen, Tang, Li, Fang, Zhu, Xie, Yin, Yi, et~al.]{wu2024vila}
Yecheng Wu, Zhuoyang Zhang, Junyu Chen, Haotian Tang, Dacheng Li, Yunhao Fang, Ligeng Zhu, Enze Xie, Hongxu Yin, Li Yi, et~al.
\newblock Vila-u: a unified foundation model integrating visual understanding and generation.
\newblock \emph{arXiv preprint arXiv:2409.04429}, 2024{\natexlab{b}}.

\bibitem[Xiang et~al.(2024)Xiang, Liu, Gu, Gao, Ning, Zha, Feng, Tao, Hao, Shi, et~al.]{xiang2024pandora}
Jiannan Xiang, Guangyi Liu, Yi Gu, Qiyue Gao, Yuting Ning, Yuheng Zha, Zeyu Feng, Tianhua Tao, Shibo Hao, Yemin Shi, et~al.
\newblock Pandora: Towards general world model with natural language actions and video states.
\newblock \emph{arXiv preprint arXiv:2406.09455}, 2024.

\bibitem[Xie et~al.(2024)Xie, Mao, Bai, Zhang, Wang, Lin, Gu, Chen, Yang, and Shou]{xie2024show}
Jinheng Xie, Weijia Mao, Zechen Bai, David~Junhao Zhang, Weihao Wang, Kevin~Qinghong Lin, Yuchao Gu, Zhijie Chen, Zhenheng Yang, and Mike~Zheng Shou.
\newblock Show-o: One single transformer to unify multimodal understanding and generation.
\newblock \emph{arXiv preprint arXiv:2408.12528}, 2024.

\bibitem[Yang et~al.(2024{\natexlab{a}})Yang, Glossop, Bhorkar, Shah, Vuong, Finn, Sadigh, and Levine]{yang2024pushing}
Jonathan Yang, Catherine Glossop, Arjun Bhorkar, Dhruv Shah, Quan Vuong, Chelsea Finn, Dorsa Sadigh, and Sergey Levine.
\newblock Pushing the limits of cross-embodiment learning for manipulation and navigation.
\newblock \emph{arXiv preprint arXiv:2402.19432}, 2024{\natexlab{a}}.

\bibitem[Yang et~al.(2023)Yang, Du, Ghasemipour, Tompson, Schuurmans, and Abbeel]{yang2023learning}
Mengjiao Yang, Yilun Du, Kamyar Ghasemipour, Jonathan Tompson, Dale Schuurmans, and Pieter Abbeel.
\newblock Learning interactive real-world simulators.
\newblock \emph{arXiv preprint arXiv:2310.06114}, 2023.

\bibitem[Yang et~al.(2024{\natexlab{b}})Yang, Walker, Parker-Holder, Du, Bruce, Barreto, Abbeel, and Schuurmans]{yang2024video}
Sherry Yang, Jacob Walker, Jack Parker-Holder, Yilun Du, Jake Bruce, Andre Barreto, Pieter Abbeel, and Dale Schuurmans.
\newblock Video as the new language for real-world decision making.
\newblock \emph{arXiv preprint arXiv:2402.17139}, 2024{\natexlab{b}}.

\bibitem[Yu et~al.(2024)Yu, Weber, Deng, Shen, Cremers, and Chen]{yu2024an}
Qihang Yu, Mark Weber, Xueqing Deng, Xiaohui Shen, Daniel Cremers, and Liang-Chieh Chen.
\newblock An image is worth 32 tokens for reconstruction and generation.
\newblock \emph{arxiv: 2406.07550}, 2024.

\bibitem[Yu et~al.(2023)Yu, Gileadi, Fu, Kirmani, Lee, Arenas, Chiang, Erez, Hasenclever, Humplik, et~al.]{yu2023language}
Wenhao Yu, Nimrod Gileadi, Chuyuan Fu, Sean Kirmani, Kuang-Huei Lee, Montse~Gonzalez Arenas, Hao-Tien~Lewis Chiang, Tom Erez, Leonard Hasenclever, Jan Humplik, et~al.
\newblock Language to rewards for robotic skill synthesis.
\newblock \emph{arXiv preprint arXiv:2306.08647}, 2023.

\bibitem[Zelikman et~al.(2022)Zelikman, Wu, Mu, and Goodman]{zelikman2022star}
Eric Zelikman, Yuhuai Wu, Jesse Mu, and Noah Goodman.
\newblock Star: Bootstrapping reasoning with reasoning.
\newblock \emph{Advances in Neural Information Processing Systems}, 35:\penalty0 15476--15488, 2022.

\bibitem[Zhang et~al.(2024)Zhang, Yin, Ye, and Gao]{zhang2024learning}
Kaifeng Zhang, Zhao-Heng Yin, Weirui Ye, and Yang Gao.
\newblock Learning manipulation skills through robot chain-of-thought with sparse failure guidance.
\newblock \emph{arXiv preprint arXiv:2405.13573}, 2024.

\bibitem[Zhao et~al.(2023)Zhao, Kumar, Levine, and Finn]{zhao23act}
Tony~Z Zhao, Vikash Kumar, Sergey Levine, and Chelsea Finn.
\newblock Learning fine-grained bimanual manipulation with low-cost hardware.
\newblock \emph{arXiv preprint arXiv:2304.13705}, 2023.

\bibitem[Zhen et~al.(2024)Zhen, Qiu, Chen, Yang, Yan, Du, Hong, and Gan]{3dvla}
Haoyu Zhen, Xiaowen Qiu, Peihao Chen, Jincheng Yang, Xin Yan, Yilun Du, Yining Hong, and Chuang Gan.
\newblock 3d-vla: A 3d vision-language-action generative world model.
\newblock \emph{arXiv preprint arXiv:2403.09631}, 2024.

\bibitem[Zhou et~al.(2024{\natexlab{a}})Zhou, Yu, Babu, Tirumala, Yasunaga, Shamis, Kahn, Ma, Zettlemoyer, and Levy]{Zhou2024TransfusionPT}
Chunting Zhou, Lili Yu, Arun Babu, Kushal Tirumala, Michihiro Yasunaga, Leonid Shamis, Jacob Kahn, Xuezhe Ma, Luke Zettlemoyer, and Omer Levy.
\newblock Transfusion: Predict the next token and diffuse images with one multi-modal model.
\newblock 2024{\natexlab{a}}.

\bibitem[Zhou et~al.(2024{\natexlab{b}})Zhou, Qin, Yin, Huang, Zhang, Sheng, Qiao, and Shao]{zhou2024minedreamer}
Enshen Zhou, Yiran Qin, Zhenfei Yin, Yuzhou Huang, Ruimao Zhang, Lu Sheng, Yu Qiao, and Jing Shao.
\newblock Minedreamer: Learning to follow instructions via chain-of-imagination for simulated-world control.
\newblock \emph{arXiv preprint arXiv:2403.12037}, 2024{\natexlab{b}}.

\bibitem[Zhou et~al.(2023)Zhou, Dean, Srirama, Rajeswaran, Pari, Hatch, Jain, Yu, Abbeel, Pinto, et~al.]{toto}
Gaoyue Zhou, Victoria Dean, Mohan~Kumar Srirama, Aravind Rajeswaran, Jyothish Pari, Kyle Hatch, Aryan Jain, Tianhe Yu, Pieter Abbeel, Lerrel Pinto, et~al.
\newblock Train offline, test online: A real robot learning benchmark.
\newblock In \emph{2023 IEEE International Conference on Robotics and Automation (ICRA)}, pages 9197--9203. IEEE, 2023.

\bibitem[Zhu et~al.(2023{\natexlab{a}})Zhu, Tian, Xu, Huo, Zhan, Tomizuka, and Ding]{zhu2023fanuc}
Xinghao Zhu, Ran Tian, Chenfeng Xu, Mingxiao Huo, Wei Zhan, Masayoshi Tomizuka, and Mingyu Ding.
\newblock Fanuc manipulation: A dataset for learning-based manipulation with fanuc mate 200id robot.
\newblock \url{https://sites.google.com/berkeley.edu/fanuc-manipulation}, 2023{\natexlab{a}}.

\bibitem[Zhu et~al.(2023{\natexlab{b}})Zhu, Joshi, Stone, and Zhu]{zhu2023viola}
Yifeng Zhu, Abhishek Joshi, Peter Stone, and Yuke Zhu.
\newblock Viola: Imitation learning for vision-based manipulation with object proposal priors.
\newblock In \emph{Conference on Robot Learning}, pages 1199--1210. PMLR, 2023{\natexlab{b}}.

\end{thebibliography}
}

\clearpage

\setcounter{page}{1}
\maketitlesupplementary

\section{Implementation Details}
\subsection{Data Details}
\label{sec:data}
We select part of the Open X-Embodiment dataset~\cite{o2023open} as our robot demonstration pre-training data, and Something2Something~\cite{goyal2017something}, and EPIC-KITCHEN-100~\cite{kapidis2019egocentric} as our action-less video data. The $u_l$ and $u_u$ is upper bound and lower bound for predicted subgoal horizon. We manually set those number for each dataset.
\begin{table}[h]
\centering
\begin{tabular}{lccc}
\hline
\textbf{Dataset} & \textbf{Weight} & $u_l$ & $u_u$ \\
\hline
Bridge~\cite{walke2023bridgedata,ebert2021bridge} & 24.14\%  & 5 & 10 \\
\hline
RT-1~\cite{brohan2022rt} & 6.90\% & 5 & 10 \\
\hline
TOTO~\cite{toto} & 10.34\% & 20 & 24 \\
\hline
VIOLA~\cite{zhu2023viola} & 10.34\% &15 & 20\\
\hline
RoboTurk~\cite{mandlekar2018roboturk} & 10.34\% &1 &2 \\
\hline
Jaco Play~\cite{taco1,taco2} & 10.34\% &10 & 15\\
\hline
Berkeley Autolab UR5~\cite{BerkeleyUR5Website} & 10.34\% &5 & 10\\
\hline
Berkeley Fanuc Manipulation~\cite{zhu2023fanuc} & 10.34\% & 10 & 15\\
\hline
Something2Something~\cite{goyal2017something} & 3.45\% & 5 & 7\\
\hline
EPIC-KITCHEN-100~\cite{kapidis2019egocentric} & 3.45\% & 5 & 7\\
\hline
\end{tabular}
\caption{Dataset Weights and Hyperparameters}
\label{tab:dataset_weights}
\end{table}

\subsection{Hyperparameters}
In this section, we list the important hyperparameters for our model pre-training and pose-training stage. 

\begin{table}[h]
\centering
\begin{tabular}{lcc}
\hline
\textbf{Hyperparameter} & \textbf{Pre-training} \\
\hline
Learning Rate & 1e-4\\
\hline 
LR Scheduler & Cosine decay \\
\hline
Global Batch Size & 2048 \\
\hline
Image Resolution & 256 × 256 \\
\hline
Action Token Size & 10\\
\hline
Epoch & 10 \\
\hline
\end{tabular}
\caption{Hyperparameters for  pre-training}
\label{tab:hyperparameters}
\end{table}
For fine-tuning on LIBERO~\cite{liu2023libero} and Franka-Tabletop~\cite{kim2024openvla} experiments, we fine-tune the model (LLM backbone, projector, depth transformer) with constant learning rate 1e-5 for 150 epochs. 

\subsection{Training}
We perform training on 12 A100 GPU nodes with 8 GPUs each. The pre-training with data mixture in~\ref{sec:data} takes 11K A100 GPU hours in total. The training cost for LIBERO and Franka-Tabletop fine-tuning is done on a single A100 GPU node for 10-24 hours depends on the dataset size. 


\end{document}